\documentclass[conference]{IEEEtran}
\IEEEoverridecommandlockouts
\usepackage{cite}
\usepackage{amsmath,amssymb,amsfonts}
\usepackage{graphicx}
\usepackage{textcomp}
\usepackage{xcolor}
\usepackage{booktabs}
\usepackage{array}
\usepackage{multirow}

\usepackage{iftex}
\ifPDFTeX
  \usepackage[T1]{fontenc}
  \usepackage[utf8]{inputenc}
  \usepackage{mathptmx} 
  \newcommand{\yo}[1]{#1}
  \newcommand{\yebelow}{\d{e}}
  \newcommand{\yobelow}{\d{o}}
  \newcommand{\ysbelow}{\d{s}}
\else
  \usepackage{fontspec}
  \newfontfamily\yorubafont{DejaVu Serif}[Scale=MatchLowercase]
  \newcommand{\yo}[1]{{\yorubafont #1}}
  \newcommand{\yebelow}{{\yorubafont ẹ}}
  \newcommand{\yobelow}{{\yorubafont ọ}}
  \newcommand{\ysbelow}{{\yorubafont ṣ}}
\fi

\newcommand{\best}[1]{\textbf{#1}}

\begin{document}

\title{Beyond Word Error Rate: A Switch-Aware Evaluation of ASR and Audio Language Models on English--Yoruba Code-Switched Speech}


\author{
  \IEEEauthorblockN{Chibuzor Okocha and Christian Grant}
  \IEEEauthorblockA{\textit{Department of Computer \& Information Science \& Engineering} \\
  \textit{University of Florida}, Gainesville, FL, USA \\
  \{c.okocha, christan\}@ufl.edu}
}

\maketitle

\begin{abstract}
Automatic speech recognition (ASR) systems and audio language models (audio LMs) now report low error rates on monolingual benchmarks, but their behavior on code-switched speech in low-resource, diacritic-rich languages remains poorly characterized. We present a switch-aware evaluation of eleven modern systems (six ASR models and five audio LMs) on English--Yoruba code-switched speech, using a deterministic 2000-utterance evaluation set and a shared scoring pipeline. Beyond word error rate (WER), we report switch-localized diagnostics: a switch-entry token error rate (SETER), windowed switch-point error rates (SPER@$k$), language-specific error rates, and a diacritic-insensitive WER. Our central finding is that aggregate WER hides code-switching behavior. The best system by WER (an ASR model) is statistically indistinguishable from a leading audio LM on WER, yet the audio LM is significantly better on every switch-localized metric. Across faithful systems, Yoruba token recognition collapses (error $\geq 0.97$ for almost all systems) while English tokens are recognized far better, and errors concentrate sharply at switches \emph{into} Yoruba. Several generative audio LMs fail as exact transcribers, producing translation, verbosity, and prompt leakage that are strongly prompt-dependent. We release manifests, metric implementations, and evaluation scripts to support reproducible, switch-aware benchmarking for African code-switched speech.
\end{abstract}


\section{Introduction}
Recent ASR systems and audio language models report strong results on widely used monolingual benchmarks. Yet a large fraction of the world's speakers are bilingual and routinely \emph{code-switch}, alternating between two languages within a single utterance. Code-switching breaks the monolingual assumptions baked into most systems: tokenizers, language priors, and decoding all degrade at language boundaries~\cite{ref_cs_survey,ref_pier}. This degradation is especially acute when one of the languages is low-resource and orthographically rich, as is the case for many African languages.

Most code-switching ASR research targets Mandarin--English (SEAME, ASCEND) or Arabic--English (ArzEn)~\cite{ref_seame,ref_ascend,ref_arzen}. English--Yoruba is comparatively underexplored, despite Yoruba being spoken by tens of millions and presenting distinctive challenges: dense lexical mixing and a tone orthography that uses sub-dot characters (\yebelow, \yobelow, \ysbelow) and tone diacritics. At the same time, evaluation practice has not kept pace with the models. Aggregate WER, the default metric, sums errors over an utterance that is usually dominated by a matrix language; errors at the rarer switch points and embedded-language tokens are diluted~\cite{ref_pier}. A model can therefore appear strong by WER while failing precisely where code-switching happens.

We take an evaluation-first stance. Rather than proposing a new model, we ask how well current systems actually transcribe English--Yoruba code-switched speech, and whether standard WER explains their behavior. We benchmark eleven systems under a shared, deterministic protocol and complement WER with switch-aware diagnostics that localize errors to switch boundaries, embedded-language tokens, and diacritics.

Our contributions are:
\begin{itemize}
\item A switch-aware benchmark of six ASR systems and five audio LMs on English--Yoruba code-switched speech, with a shared, deterministic 2000-utterance evaluation set and paired statistical testing.
\item Evidence that WER and switch-localized metrics rank systems differently: the best ASR model by WER is statistically tied with a leading audio LM that is in turn significantly better at switch boundaries. Across systems, WER rank is essentially uncorrelated with Yoruba token error and only weakly correlated with switch-entry error, so WER is a poor proxy for code-switching fidelity.
\item A characterization of Yoruba recognition collapse and a strong, system-consistent switch-direction asymmetry, together with the effects of switch density, language dominance, and domain.
\item An analysis of generative audio LM failure modes (translation, verbosity, prompt leakage), their prompt sensitivity, and an automatic error taxonomy that separates recognition from generation errors without human labels.
\item Released manifests, metric implementations, and evaluation scripts for reproducible, switch-aware benchmarking of African code-switched speech.
\end{itemize}

\section{Related Work}
\textbf{Code-switching ASR and metrics.} Code-switching ASR has been studied mainly on Mandarin--English (SEAME, ASCEND) and Arabic--English (ArzEn) corpora~\cite{ref_seame,ref_ascend,ref_arzen}, typically scored with WER, CER, and mixed error rate (MER), which counts errors per token in each language but still aggregates over the whole utterance. Because the matrix language dominates most utterances, such aggregate metrics under-weight the embedded-language tokens and switch points that distinguish code-switching from monolingual ASR. Ugan et al.~\cite{ref_pier} make this explicit: fine-tuning on monolingual data from both languages \emph{improves} classical WER on code-switched test sets even as accuracy on the actually code-switched words degrades. They propose the Point-of-Interest Error Rate (PIER), which restricts scoring to embedded-language tokens, and the DECM benchmark for bilingual ASR~\cite{ref_decm}. Related diagnostics include switch-point WER and language-pair-specific error rates. Our diagnostics are complementary to PIER: instead of scoring by language membership alone, SETER and SPER@$k$ localize scoring to the switch \emph{boundary} (the first embedded token and a window around it), and we additionally report per-language error rates and a diacritic-insensitive WER. To our knowledge, these switch-localized diagnostics have not previously been applied to English--Yoruba or, more broadly, to a side-by-side comparison of ASR systems and audio LMs.

\textbf{Audio language models.} Instruction-following audio LMs couple a speech encoder to a large language model and are prompted in natural language. Recent systems include Qwen2-Audio and the omni-modal Qwen2.5-Omni~\cite{ref_qwen2audio,ref_qwenomni}, Kimi-Audio~\cite{ref_kimi}, and the Audio Flamingo family~\cite{ref_af2,ref_af3}; LLM-decoder ASR models such as SALM~\cite{ref_salm} (the basis of Canary-Qwen) blur the line further. These models are usually benchmarked on audio \emph{understanding} and question-answering tasks, where free-form generation is desirable; their reliability as faithful, verbatim transcribers of code-switched, low-resource speech is much less examined, and the generation-oriented decoding that helps understanding can hurt transcription. Concurrent work has begun to probe audio LMs' semantic reasoning over African-accented speech~\cite{ref_afrisem}; we focus instead on verbatim transcription of code-switched speech. We therefore evaluate them head-to-head with dedicated ASR systems spanning transducer (Parakeet-TDT~\cite{ref_nemo,ref_tdt}), attention-encoder-decoder (Whisper~\cite{ref_whisper}), and LLM-decoder (Canary-Qwen~\cite{ref_salm}, Qwen3-ASR~\cite{ref_qwen3asr}, Granite Speech~\cite{ref_granite}) designs.

\textbf{Low-resource and African speech.} African languages remain under-served by speech technology despite large speaker populations. A growing body of resources and benchmarks targets African-accented English and African languages, including AfriSpeech-200~\cite{ref_afrispeech} for clinical and general-domain ASR, AfriSpeech-Dialog~\cite{ref_afridialog} for spontaneous conversation, AfriSpeech-MultiBench~\cite{ref_afrimulti} and AfriVox~\cite{ref_afrivox} for multidomain and multilingual evaluation of ASR systems and speech LLMs, and AfriNames~\cite{ref_afrinames} for entity-rich recognition. Yoruba in particular has motivated work on diacritic (tone and sub-dot) restoration~\cite{ref_yoruba}, since stripped diacritics are common in web text and change word identity. Yet most of these resources target monolingual ASR, spoken-language understanding, or text-to-speech; code-switched, diacritic-aware \emph{evaluation} of modern ASR systems and audio LMs is largely absent. We address this gap for English--Yoruba and treat diacritics explicitly through a diacritic-insensitive WER variant.

\section{Corpus}
We evaluate on \textsc{AfriCodeSwitch}, an English--Yoruba code-switched speech corpus. We use the validation partition and stage a deterministic set of evaluation manifests so that every system is scored on identical utterances. Table~\ref{tab:corpus} summarizes the audited statistics. Of 9{,}966 metadata rows, 8{,}181 have available audio (9.54 hours); the remaining 1{,}785 rows lack audio and are excluded. Utterances are short (mean 4.20\,s, 8.52 words) and densely mixed (mean 2.60 switches, up to 9), spanning 100 speakers, 2{,}937 prompts, and 13 domains.

Word-level language tags are parsed from the corpus metadata and used to derive, for each utterance, switch points, switch direction, language dominance (English-dominant, balanced, Yoruba-dominant), and switch density. Text is Unicode NFC-normalized. The corpus is balanced across dominance (2{,}799 English-dominant, 3{,}336 balanced, 2{,}046 Yoruba-dominant) but speaker gender is skewed (6{,}969 female vs.\ 1{,}212 male rows), which we note as a limitation. Unless stated otherwise, results use the 2000-utterance \texttt{cap2000} manifest; prompt-robustness studies use \texttt{cap1000}.

\begin{table}[t]
\caption{Audited statistics of the \textsc{AfriCodeSwitch} validation corpus.}
\label{tab:corpus}
\centering
\small
\begin{tabular}{lr}
\toprule
Statistic & Value \\
\midrule
Metadata rows & 9{,}966 \\
Rows with available audio & 8{,}181 \\
Rows missing audio & 1{,}785 \\
Available audio (hours) & 9.54 \\
Speakers / prompts / domains & 100 / 2{,}937 / 13 \\
Mean duration (s) & 4.20 \\
Mean words / utterance & 8.52 \\
Mean / max switches per utterance & 2.60 / 9 \\
Dominance (EN / bal. / YO) & 2{,}799 / 3{,}336 / 2{,}046 \\
\bottomrule
\end{tabular}
\end{table}

\section{Switch-Aware Metrics}
All metrics derive from a minimum-edit-distance alignment between the reference token sequence $R=(r_1,\dots,r_N)$ and the hypothesis. The alignment yields substitution, deletion, and insertion counts $S,D,I$ and labels each reference token as correctly recognized or not. Let $e_i=1$ if $r_i$ is substituted or deleted (not correctly recognized) and $e_i=0$ otherwise. For any set of reference positions $T\subseteq\{1,\dots,N\}$ we define the \emph{masked error rate}
\begin{equation}
\mathrm{ER}(T)=\frac{1}{|T|}\sum_{i\in T} e_i .
\label{eq:er}
\end{equation}
Every language-specific and switch-aware metric below is an instance of \eqref{eq:er} over a different position set; corpus-level values are the mean over utterances.

\textbf{WER and CER.} Word error rate is the standard
\begin{equation}
\mathrm{WER}=\frac{S+D+I}{N},
\end{equation}
computed after NFC normalization, case-folding, and punctuation stripping; CER is the analogous ratio over characters. Normalization is diacritic-sensitive by default.

\textbf{Diacritic-insensitive WER.} Let $\phi(\cdot)$ strip combining tone and sub-dot marks from a token. $\mathrm{WER}_{\mathrm{di}}$ is WER recomputed on $\phi(R)$ and $\phi(H)$; comparing it to WER isolates the error attributable to Yoruba diacritics.

\textbf{Language-specific error rates.} With $\mathrm{lang}(r_i)\in\{\textsc{en},\textsc{yo}\}$ taken from the reference tags,
\begin{equation}
\text{EN-ER}=\mathrm{ER}(L_{\textsc{en}}),\quad \text{YO-ER}=\mathrm{ER}(L_{\textsc{yo}}),
\end{equation}
where $L_{\ell}=\{i:\mathrm{lang}(r_i)=\ell\}$. These quantify which language a system fails on.

\textbf{SETER (Switch-Entry Token Error Rate).} A switch-entry token is the first token of a new language run. Defining the switch-entry set
\begin{equation}
P=\{\,i\ge 2 : \mathrm{lang}(r_i)\neq\mathrm{lang}(r_{i-1})\,\},
\end{equation}
SETER$=\mathrm{ER}(P)$ measures whether a system ``lands'' each switch.

\textbf{SPER@$k$ (Switch-Point Error Rate).} Let $d_i=\min_{j\in P}|i-j|$ be the distance from position $i$ to the nearest switch entry, and let $W_k=\{i:d_i\le k\}$ be the window of radius $k$ around switch points. Then
\begin{equation}
\text{SPER@}k=\mathrm{ER}(W_k),\qquad k\in\{0,1,2,3\}.
\end{equation}
By construction $\text{SPER@}0=\text{SETER}$ (the entry token alone), and increasing $k$ widens the neighbourhood toward the whole utterance. We foreground SPER@1 and SPER@3.

\textbf{Output-quality flags.} Per-utterance heuristics flag blank output, truncation, degenerate repetition, length expansion (hypothesis far longer than reference), and prompt leakage (instruction text echoed in the output); they characterize generative failure modes rather than scoring transcription accuracy.

Relative to PIER~\cite{ref_pier}, which scores embedded-language ``points of interest,'' SETER and SPER@$k$ are positional diagnostics centred on switch boundaries, while the language-specific rates capture the complementary language-membership view.\footnote{Exact normalization, alignment, and window conventions follow our released implementation.}

\section{Experimental Setup}
We evaluate six ASR systems: \textit{Parakeet-TDT-0.6B-v2}~\cite{ref_nemo,ref_tdt}, \textit{Qwen3-ASR-1.7B}~\cite{ref_qwen3asr}, \textit{Canary-Qwen-2.5B}~\cite{ref_nemo,ref_salm}, \textit{Whisper-large-v3}~\cite{ref_whisper} (forced English and auto language detection), and \textit{Granite-Speech-4.1-2B}~\cite{ref_granite}; and five audio LMs: \textit{Kimi-Audio-7B-Instruct}~\cite{ref_kimi}, \textit{Audio Flamingo~3}~\cite{ref_af3}, \textit{Audio Flamingo~2}~\cite{ref_af2}, \textit{Qwen2-Audio-7B-Instruct}~\cite{ref_qwen2audio}, and \textit{Qwen2.5-Omni-7B}~\cite{ref_qwenomni}. All systems run zero-shot with greedy decoding. The main benchmark uses a single transcription instruction (the \emph{primary} prompt) for all audio LMs. For two prompt-sensitive audio LMs we additionally evaluate \emph{direct} and \emph{anti-translation} prompt variants. Inference and scoring are driven by manifest-based runners on a Slurm cluster; the same manifests, normalization, and alignment are used for every system. Two intended systems (Qwen3-Omni and a gated Cohere transcription model) could not be run due to environment and access constraints and are left to future work.

\section{Results}

\begin{table*}[t]
\caption{Main \texttt{cap2000} benchmark (2{,}000 shared utterances, primary prompt). All values are percentages; lower is better. WER$_\mathrm{di}$: diacritic-insensitive WER; EN/YO-ER: English/Yoruba token error rate. Best value per column among faithful systems is in \best{bold}. The lower block contains generative models whose WER${>}100\%$ indicates substantial over-generation rather than ordinary substitution.}
\label{tab:main}
\centering
\small
\begin{tabular}{l l rrr rr rrr}
\toprule
Model & Family & WER & CER & WER$_\mathrm{di}$ & EN-ER & YO-ER & SETER & SPER@1 & SPER@3 \\
\midrule
Parakeet-TDT-0.6B-v2      & ASR  & \best{66.1} & \best{35.7} & 66.0 & 34.4 & 99.0 & 68.9 & 67.2 & 65.2 \\
Kimi-Audio-7B-Instruct    & LALM & 66.2 & 37.4 & \best{64.3} & \best{32.9} & 97.5 & \best{66.8} & \best{65.0} & \best{63.2} \\
Qwen3-ASR-1.7B            & ASR  & 67.6 & 39.6 & 67.9 & 37.9 & 99.2 & 70.4 & 68.2 & 66.0 \\
Canary-Qwen-2.5B          & ASR  & 70.0 & 40.1 & 70.0 & 33.5 & 99.4 & 68.7 & 66.8 & 64.8 \\
Audio Flamingo 3          & LALM & 70.1 & 38.9 & 69.5 & 40.2 & 98.4 & 73.1 & 69.7 & 67.3 \\
Whisper-large-v3          & ASR  & 71.2 & 41.7 & 70.9 & 36.8 & 98.9 & 70.2 & 68.0 & 66.0 \\
Granite-Speech-4.1-2B     & ASR  & 73.5 & 42.8 & 74.0 & 38.9 & 99.4 & 72.9 & 69.6 & 67.2 \\
Whisper-large-v3 (auto)   & ASR  & 74.8 & 39.2 & 71.7 & 46.0 & \best{96.6} & 73.0 & 71.3 & 69.2 \\
\midrule
Audio Flamingo 2          & LALM & 119.8 & 112.3 & 128.8 & 98.0 & 99.7 & 99.2 & 99.3 & 98.9 \\
Qwen2-Audio-7B-Instruct   & LALM & 137.1 & 123.4 & 146.3 & 57.7 & 98.8 & 77.2 & 76.4 & 75.3 \\
Qwen2.5-Omni-7B           & LALM & 202.9 & 123.0 & 143.4 & 57.3 & 91.3 & 78.0 & 75.2 & 72.8 \\
\bottomrule
\end{tabular}
\end{table*}

\subsection{WER does not explain switch behavior}
Table~\ref{tab:main} reports the main benchmark. Parakeet attains the best WER (66.1\%) and CER (35.7\%). Strikingly, Kimi-Audio, a generative audio LM, ties it on WER (66.2\%) while winning five of the remaining seven columns, including the diacritic-insensitive WER, the English error rate, and all three switch-localized metrics (SETER, SPER@1, SPER@3). In other words, the two systems that look identical by aggregate WER behave differently exactly where code-switching occurs.

A paired utterance-level bootstrap (1{,}000 resamples over the shared manifest) confirms this dissociation (Table~\ref{tab:sig}). The WER difference between Parakeet and Kimi is not significant ($\Delta=+0.13$ points, $p=0.76$, CI crossing zero), whereas Kimi's advantage on SETER, SPER@1, and SPER@3 is significant ($p<0.001$ in all three cases). The effect is not unique to this pair: Canary-Qwen has significantly worse WER than Parakeet ($p<0.001$) but statistically indistinguishable SETER and SPER ($p>0.12$). Switch-localized fidelity and aggregate WER are therefore separable axes of performance.

This separation holds across the whole pool. Ranking the systems by WER and by each diagnostic, the Spearman correlation between WER and English token error is high ($\rho=0.74$), but the correlation between WER and Yoruba token error is essentially zero ($\rho=-0.10$), and the correlation between WER and switch-entry error (SETER) is only moderate ($\rho=0.67$, not significant at $\alpha=0.05$). A system's aggregate WER is thus largely determined by its matrix-language (English) accuracy and tells us little about the code-switching-specific behavior the task cares about. The boundary window itself is the hardest region: SPER@0 (the switch token) exceeds SPER@3 for every system (e.g., $68.9\%\!\rightarrow\!65.2\%$ for Parakeet), so error eases only gradually as the window widens away from the switch.

\begin{table}[t]
\caption{Paired bootstrap comparison, Kimi-Audio vs.\ Parakeet (\texttt{cap2000}). $\Delta$ is Kimi minus Parakeet in points; negative favors Kimi. 95\% CIs and two-sided bootstrap $p$.}
\label{tab:sig}
\centering
\small
\begin{tabular}{l r r l}
\toprule
Metric & $\Delta$ (pts) & 95\% CI & $p$ \\
\midrule
WER     & $+0.13$ & $[-0.67, +1.12]$ & $0.76$ \\
SETER   & $-2.09$ & $[-3.33, -0.83]$ & $<0.001$ \\
SPER@1  & $-2.20$ & $[-3.01, -1.38]$ & $<0.001$ \\
SPER@3  & $-2.02$ & $[-2.64, -1.37]$ & $<0.001$ \\
\bottomrule
\end{tabular}
\end{table}

\subsection{Yoruba recognition collapses}
Errors are overwhelmingly driven by Yoruba, not by ordinary English ASR error. Among the eight faithful systems, English token error ranges from 33\% to 46\%, while Yoruba token error is $\geq 96.6\%$ and is $\geq 97.5\%$ for every system other than auto-language Whisper. In effect, English tokens are recognized roughly two-thirds of the time, while Yoruba tokens are almost never correct. We recomputed the language-specific rates directly from manifest tags and alignments as a sanity check; the recomputed values match the reported rates exactly for the top systems (e.g., Parakeet YO-ER $0.9903$, EN-ER $0.3444$), confirming the collapse is real and not a scoring artifact.

\subsection{Errors localize at switches into Yoruba}
\begin{figure*}[t]
\centering
\includegraphics[width=\textwidth]{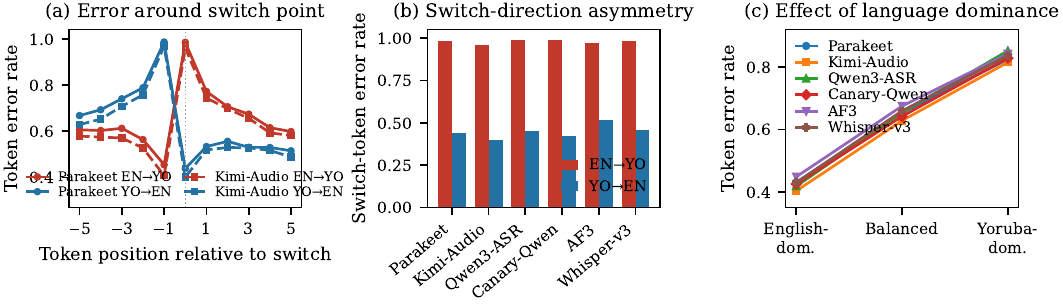}
\caption{Structural error analysis on \texttt{cap2000}. (a) Token error vs.\ position relative to the switch point for Parakeet and Kimi-Audio, by direction; error peaks at switches into Yoruba. (b) Switch-token error by direction across strong systems: switching into Yoruba is near-uniformly hard. (c) Token error increases monotonically with Yoruba dominance. EN$\rightarrow$YO: switch into Yoruba; YO$\rightarrow$EN: switch into English.}
\label{fig:struct}
\end{figure*}

Figure~\ref{fig:struct} decomposes errors structurally. Panel (a) plots token error against position relative to a switch: error spikes at the boundary when the switch is \emph{into} Yoruba, but stays far lower when the switch is into English. Panel (b) makes the asymmetry explicit: switch-token error for EN$\rightarrow$YO is $0.95$--$1.00$ across systems, versus $0.40$--$0.53$ for YO$\rightarrow$EN. Panel (c) shows a monotonic effect of dominance: error climbs from $\sim$0.42 on English-dominant utterances to $\sim$0.65 (balanced) to $\sim$0.83 (Yoruba-dominant). Kimi-Audio is consistently lowest within each stratum. These patterns indicate that switch-boundary degradation is a distinct failure mode tied to producing Yoruba content, not a uniform smearing of WER.

\subsection{Switching rate, domain, and speaker}
\begin{figure}[t]
\centering
\includegraphics[width=\columnwidth]{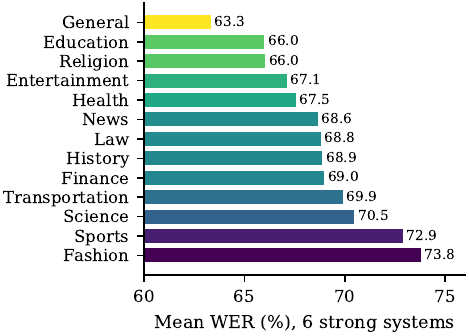}
\caption{Mean WER per domain on \texttt{cap2000}, averaged over the six strong systems. Difficulty spans a 10-point range; terminology-dense domains (Fashion, Sports, Science) are hardest.}
\label{fig:domain}
\end{figure}

Three further factors modulate difficulty. \textbf{Switching rate:} grouping utterances by switch density, mean WER over the strong systems rises monotonically from $66.9\%$ (low) to $67.9\%$ (moderate) to $71.2\%$ (high), and SETER rises more steeply ($66.7\%\!\rightarrow\!70.1\%\!\rightarrow\!72.1\%$): denser switching hurts boundary fidelity more than it hurts aggregate WER. \textbf{Domain:} difficulty is heterogeneous across the 13 domains (Fig.~\ref{fig:domain}), spanning a $10$-point WER range from conversational/expository domains (General, Education, Religion; $\sim$63--66\%) to terminology-heavy ones (Science, Sports, Fashion; $\sim$71--74\%), reflecting how much rare, often borrowed vocabulary each domain carries. \textbf{Speaker gender:} despite the corpus skew toward female speakers, error rates are comparable across gender (mean WER $68.9\%$ male vs.\ $68.4\%$ female; Yoruba error $0.99$ vs.\ $0.99$). The dominant disparity in this benchmark is along the language axis, not speaker gender, a point we return to in Section~\ref{sec:ethics}.

\subsection{Diacritics are not yet the bottleneck}
Stripping diacritics barely changes WER for the strong systems (Parakeet $+0.06$, Canary $-0.05$, Qwen3-ASR $-0.29$, Whisper-v3 $+0.30$ points). The exception among faithful systems is Kimi-Audio ($+1.85$ points), consistent with it producing more Yoruba content, and therefore more diacritics to get wrong. Because most systems rarely produce correct Yoruba at all, diacritic accuracy is currently a second-order concern; it will become central once base Yoruba recognition improves. We report WER$_\mathrm{di}$ so that future, stronger systems can be assessed on diacritics directly.

\subsection{Generative audio LMs often fail as transcribers}
The lower block of Table~\ref{tab:main} shows three audio LMs with WER above 100\%, indicating systematic over-generation rather than substitution. Output-quality flags localize the cause: Qwen2-Audio leaks the prompt into the transcript in 32\% of \texttt{cap2000} outputs, and Qwen2.5-Omni shows elevated repetition (3.1\%) and length expansion (3.7\%), including degenerate loops. By contrast, all ASR systems and Kimi-Audio are clean on these flags. These behaviors are highly prompt-dependent (Table~\ref{tab:prompt}): switching Qwen2-Audio from the primary prompt to a direct instruction cuts prompt leakage from 33.7\% to 0.2\% and WER from 135.6\% to 103.3\%; Qwen2.5-Omni improves similarly. Prompting is thus a major confound when benchmarking generative audio LMs for transcription, and a single prompt can under-state or over-state a model's true transcription ability.

\subsection{An automatic error taxonomy}
Many failure modes are detectable directly from the alignment and the output, without human labeling. Table~\ref{tab:taxonomy} summarizes two such views. The first decomposes word-level errors into substitutions, deletions, and insertions. Every faithful system is substitution-dominated (69--73\% of errors), the signature of \emph{recognition} failure: the model emits the wrong word rather than inventing or dropping content. The two worst audio LMs invert this profile: insertions account for 46\% (Qwen2-Audio) and 65\% (Qwen2.5-Omni) of their errors, the signature of \emph{generation} failure. The second view counts generative pathologies per utterance. These are negligible for ASR systems and Kimi-Audio but isolate the specific breakdown of each weak model: prompt leakage dominates Qwen2-Audio (31.9\%), while repetition loops and length expansion dominate Qwen2.5-Omni. This automatic taxonomy cleanly separates recognition from generation errors, but it cannot by itself distinguish a faithful mis-transcription from a fluent translation or paraphrase, which requires output-side language identification or human judgment. The curated 100-utterance sample (below) targets exactly those semantic categories.

\begin{table}[t]
\caption{Automatic error taxonomy on \texttt{cap2000} (primary). Left: composition of word-level errors into substitutions/deletions/insertions (\% of errors). Right: per-utterance incidence (\%) of generative pathologies: prompt leakage, repetition, length expansion. Recognition-error systems are substitution-dominated; over-generating systems are insertion-dominated.}
\label{tab:taxonomy}
\centering
\footnotesize
\setlength{\tabcolsep}{4pt}
\begin{tabular}{l rrr rrr}
\toprule
& \multicolumn{3}{c}{Error composition} & \multicolumn{3}{c}{Pathology rate} \\
\cmidrule(lr){2-4}\cmidrule(lr){5-7}
Model & Sub & Del & Ins & Leak & Rep & Exp \\
\midrule
Parakeet      & 69.2 & 26.9 & 3.8  & 0.2 & 0.1 & 0.0 \\
Kimi-Audio    & 70.1 & 22.8 & 7.1  & 0.3 & 0.2 & 0.1 \\
Qwen3-ASR     & 71.0 & 23.7 & 5.3  & 0.2 & 0.1 & 0.1 \\
Canary-Qwen   & 73.2 & 17.0 & 9.8  & 0.2 & 0.2 & 0.1 \\
Audio Flamingo 3 & 72.9 & 20.4 & 6.7 & 0.2 & 0.5 & 0.2 \\
Whisper-v3    & 69.5 & 20.6 & 9.9  & 0.2 & 0.3 & 0.1 \\
Granite       & 71.5 & 17.4 & 11.2 & 0.3 & 0.2 & 0.1 \\
Whisper-auto  & 72.2 & 17.6 & 10.2 & 0.2 & 0.7 & 0.4 \\
\midrule
Audio Flamingo 2 & 73.3 & 8.8 & 17.8 & 0.2 & 0.8 & 1.3 \\
Qwen2-Audio   & 52.1 & 1.6  & 46.2 & \best{31.9} & 0.9 & 3.5 \\
Qwen2.5-Omni  & 30.8 & 4.3  & 64.9 & 0.7 & \best{3.1} & \best{3.7} \\
\bottomrule
\end{tabular}
\end{table}

\begin{table}[t]
\caption{Prompt robustness on \texttt{cap1000}. Lower WER/SETER is better; ``Leak'' is the prompt-leakage rate.}
\label{tab:prompt}
\centering
\small
\begin{tabular}{l l rrr}
\toprule
Model & Prompt & WER & SETER & Leak \\
\midrule
\multirow{3}{*}{Qwen2-Audio-7B} & primary          & 135.6 & 76.5 & 33.7 \\
                                & direct           & \best{103.3} & 77.6 & \best{0.2} \\
                                & anti-translation & 139.6 & 81.0 & 0.3 \\
\midrule
\multirow{3}{*}{Qwen2.5-Omni-7B} & primary          & 183.0 & 77.6 & 0.7 \\
                                 & direct           & \best{125.9} & 75.8 & 1.6 \\
                                 & anti-translation & 194.1 & 77.3 & 0.4 \\
\bottomrule
\end{tabular}
\end{table}

\section{Qualitative Analysis}
The metrics above are illustrated by recurring patterns in the outputs. We highlight three.

\textbf{Landing the switch.} For \yo{Ó bá wa nu} \textit{chalkboard} (one switch), Kimi-Audio recovers the embedded word (``\ldots nu chalkboard'', SETER $=0$) while Parakeet, Qwen3-ASR, Canary, Audio Flamingo~3, and Whisper all miss it (``chopboard'', ``chobod'', ``chopper'', \ldots; SETER $=1$). For \textit{mo love airports, but mo hate flying}, Kimi-Audio and Whisper transcribe it verbatim (WER $=0$), whereas Parakeet hallucinates English (``Molov Airport Bomo eight flying'').

\textbf{WER wins, switch lost.} The advantage is not universal. On a Yoruba-heavy utterance, Kimi-Audio enters a degenerate loop (repeating a short phrase, WER $=4.39$) while Parakeet returns a short, wrong-but-bounded hypothesis (WER $=1.00$), a case where Parakeet's lower WER masks that neither system recovers the switch.

\textbf{Translation and language confusion.} Generative models frequently translate or summarize instead of transcribing (e.g., rendering a Yoruba clause as an English paraphrase), and occasionally emit text in unrelated scripts. These are exactly the behaviors penalized by switch-localized metrics but partially hidden by utterance-level WER on matrix-heavy inputs.

Complementing the automatic taxonomy of Table~\ref{tab:taxonomy}, we curated a 100-utterance sample across the top five systems for fine-grained human annotation of the categories that automation cannot resolve: in particular translation-instead-of-transcription, along with Yoruba deletion/substitution, English hallucination, boundary collapse, and diacritic loss. The human-labeled taxonomy is in progress and released with the benchmark.

\section{Discussion}
\textbf{Why WER hides switching here.} The dissociation we observe is structural, not incidental. In an asymmetric code-switching setting the matrix language (English) supplies most reference tokens and is recognized two to three times more accurately than the embedded language, so aggregate WER is dominated by matrix-language accuracy, exactly what the cross-system correlations show (WER tracks EN-ER at $\rho=0.74$ but is uncorrelated with YO-ER at $\rho=-0.10$). A single corpus-level WER therefore cannot, even in principle, reflect how well a system handles the embedded language or the switch points that define the task. For low-resource code-switching we accordingly argue that language-specific and switch-localized metrics should be reported as standard alongside WER, not treated as optional diagnostics.

\textbf{The bottleneck is Yoruba coverage, not diacritics.} Two findings together locate the failure: Yoruba token error sits near ceiling for every faithful system, yet stripping diacritics barely moves WER. Models are thus not getting Yoruba words almost-right modulo tone marks; they largely fail to produce Yoruba content at all. This implies an ordering of priorities for practitioners: expanding Yoruba lexical and acoustic coverage (data, tokenization, pretraining exposure) must precede orthographic refinements such as diacritic restoration, which only become measurable once base recognition improves. We report a diacritic-insensitive WER precisely so this second-order effect can be tracked as systems get better.

\textbf{Switch direction points to a matrix-language prior.} The strong asymmetry (near-total error when switching \emph{into} Yoruba but roughly half that when switching back into English) suggests that decoding is governed by a dominant high-resource prior: models readily snap back to English after an embedded span but resist entering Yoruba. This is consistent with boundary errors arising from language balance within the model rather than from acoustic difficulty alone, and it motivates interventions that act at the boundary specifically: language-balanced or switch-aware decoding, and supervision targeted at embedded-language entry tokens rather than at uniform WER.

\textbf{Audio LMs cut both ways.} Kimi-Audio shows that an instruction-following audio LM can match the best dedicated ASR system on WER while significantly outperforming it at switch boundaries, plausibly because an LLM decoder carries broader multilingual text knowledge that helps it realize Yoruba spans. Yet the same model family also contains the worst transcribers in our pool, failing through translation, verbosity, and prompt leakage rather than mis-recognition; these pathologies are largely artifacts of prompting and decoding, since a direct prompt cut Qwen2-Audio's leakage from $34\%$ to near zero. Audio LMs are thus a promising route to better embedded-language recognition, but only when constrained toward verbatim transcription, and any fair benchmark of them must control the prompt. More broadly, our results caution against ranking code-switching systems by aggregate WER, a practice with direct fairness consequences: WER-only leaderboards render the exclusion of embedded-language speakers invisible. They also chart concrete next steps: Yoruba- and boundary-focused fine-tuning, language-balanced decoding, and transcription-constrained prompting for audio LMs.

\section{Limitations and Ethics}
\label{sec:ethics}
This is an evaluation-only study on a single English--Yoruba corpus; no training or fine-tuning is performed, and results may not transfer to other dialects, accents, or recording conditions. The corpus is gender-skewed toward female speakers; reassuringly, we find no large error gap between male and female speakers in this sample, which suggests the dominant disparity is along the language axis (Yoruba vs.\ English) rather than speaker demographics, though the skew still limits the strength of any per-gender claim. Switch-aware metrics are computed from reference-side language tags and an automatic alignment; tag and alignment quality bound their precision, and we mitigate this with a recomputation sanity check. The fine-grained error taxonomy relies on human annotation that is still underway. We view low resource code-switched evaluation as an accessibility and fairness concern: systems that cannot transcribe Yoruba effectively exclude its speakers, and reporting only WER can make this exclusion invisible.

\section{Conclusion}
We presented a switch-aware evaluation of modern ASR systems and audio language models on English--Yoruba code-switched speech, showing that aggregate WER masks a distinct, boundary-localized failure mode that switch-localized and language-specific metrics bring into view. We release the manifests, metric implementations, and evaluation scripts to make these failures visible and to provide a reproducible baseline for African code-switched speech; switch-aware fine-tuning and adaptation are natural next steps.

\section*{Acknowledgment}
The authors thank the contributors and annotators who made the \textsc{AfriCodeSwitch} corpus possible, and gratefully acknowledge the computational resources that supported these experiments. We also thank colleagues for their feedback on earlier versions of this work.

\end{document}